\documentclass[11pt]{article} 
\usepackage{rldmsubmit,palatino}
\usepackage{graphicx}
\usepackage{amsmath}
\usepackage{amssymb}
\usepackage{cite}
\usepackage{cleveref}

\usepackage[compact]{titlesec}         
\titlespacing{\section}{0pt}{0pt}{0pt} 
\AtBeginDocument{
  \setlength\abovedisplayskip{0pt}
  \setlength\belowdisplayskip{0pt}}

\DeclareMathOperator*{\truemax}{max} 

\newcommand{\PHUMES}{\textsc{Phumes }}
\newcommand{\PLUMES}{\textsc{Plumes }}

\title{Robotic Planning under Uncertainty in Spatiotemporal Environments for Expeditionary Science}

\author{
Victoria~Preston*\\
MIT-WHOI Joint Program\\
Massachusetts Institute of Technology\\
Cambridge, MA 02139 \\
\texttt{vpreston@csail.mit.edu} \\
\And
Genevieve~Flaspohler*\\
MIT-WHOI Joint Program\\
Massachusetts Institute of Technology\\
Cambridge, MA 02139 \\
\texttt{geflaspo@csail.mit.edu} \\
\AND
Anna P.~M.~Michel \\
Applied Ocean Physics and Engineering \\
Woods Hole Oceanographic Institution \\
\texttt{amichel@whoi.edu} \\
\And
John W.~Fisher III \\
Computer Science and Artificial Intelligence Laboratory\\
Massachusetts Institute of Technology\\
Cambridge, MA 02139 \\
\texttt{fisher@csail.mit.edu} \\
\And
Nicholas~Roy \\
Computer Science and Artificial Intelligence Laboratory\\
Massachusetts Institute of Technology\\
Cambridge, MA 02139 \\
\texttt{nickroy@csail.mit.edu} \\
}

\begin{document}

\maketitle

\begin{abstract}
In the expeditionary sciences, spatiotemporally varying environments --- hydrothermal plumes, algal blooms, lava flows, or animal migrations --- are ubiquitous. Mobile robots are uniquely well-suited to study these dynamic, mesoscale natural environments. 
We formalize expeditionary science as a \emph{sequential decision-making problem}, modeled using the language of partially-observable Markov decision processes (POMDPs). Solving the expeditionary science POMDP under real-world constraints requires efficient probabilistic modeling and decision-making in problems with complex dynamics and observational models. Previous work in informative path planning, adaptive sampling, and experimental design have shown compelling results, largely in static environments, using data-driven models and information-based rewards. However, these methodologies do not trivially extend to expeditionary science in spatiotemporal environments: they generally do not make use of scientific knowledge such as equations of state dynamics, they focus on information gathering as opposed to scientific task execution, and they make use of decision-making approaches that scale poorly to large, continuous problems with long planning horizons and real-time operational constraints. In this work, we discuss these and other challenges related to probabilistic modeling and decision-making in expeditionary science, and present some of our preliminary work that addresses these gaps. We ground our results in a real expeditionary science deployment of an autonomous underwater vehicle (AUV) in the deep ocean for hydrothermal vent discovery and characterization. Our concluding thoughts highlight remaining work to be done, and the challenges that merit consideration by the reinforcement learning and decision-making community.
\end{abstract}


\acknowledgements{Support for scientific cruise RR2107 was supported by NSF OCE OTIC \#1842053 and we thank the Captain and Crew of R/V Revelle, AUV SENTRY and ROV JASON teams, Scott Wankel, Peter Girguis, and colleagues for cruise assistance. V.P. is funded by a Martin Fellowship and in part by a NDSEG Fellowship. G.F. is funded by a Microsoft Research Fellowship and in part by award DE-NA000392 under the Department of Energy/National Nuclear Security Administration.}

\startmain 
\section{Introduction}

Expeditionary science is the act of collecting \emph{in situ} samples and observations of the natural world for the purposes of scientific discovery and model development. Transient, dynamic phenomena --- hydrothermal plumes, algal blooms, lava flows --- are of interest in many disciplines of observational science. It is often impossible for a single, static sensor to capture a comprehensive picture of these mesoscale spatiotemporal phenomena due to their large spatial scales and dynamic temporal nature. Large networks of static sensor nodes could address this sensing challenge, but the transience and sparsity of the target phenomena necessitates logistically impractical sensor density. Mobile robots are uniquely well-positioned to tackle this problem via extended deployments carrying heterogeneous sensor payloads that actively seek and perform targeted surveys of ephemeral environments. However, robots deployed for expeditionary science today currently execute open-loop, preset trajectories, like ``lawnmowers'' (back-and-forth grid-based pattern) hand-designed by human scientists. In dynamic environments, this open-loop execution often results in sparse measurements of the target phenomena or misses a short-lived target entirely. Given the cost of expeditionary field operations and the value of the data collected, it is critical to improve the efficiency and efficacy of robots as scientific tools.


Building robotic platforms for expeditionary science requires an autonomy stack that can integrate observations from heterogeneous sensors into a model of a spatiotemporal system and use this model to plan informative trajectories that target a specific scientific objective. This kind of autonomy poses many challenges for integrating probabilistic modeling and decision-making. Previous work in informative path planning (IPP) \cite{Hitz2017}, adaptive sampling/experimental design \cite{krause2008near}, and decision-making under uncertainty \cite{sunberg2018online} has tackled aspects of the expeditionary science problem, especially in static environments using data-driven models and information-based rewards. However, existing methodologies do not trivially extend to spatiotemporal environments, leaving key challenges such as model and dynamics learning and decision-making in large-scale environments unaddressed. In this abstract, we discuss these and other key challenges, and present some preliminary work conducted for hydrothermal plume discovery and mapping in the deep sea. 

\section{Expeditionary Science as a Decision-Making Problem}
In an expeditionary science mission, a robot is tasked with collecting scientifically useful measurements of an unknown, partially-observable spatiotemporal environment under time, energy, and dynamical constraints. We formulate this sequential decision-making problem as a partially observable Markov decision-process (POMDP). Let $\Pi(\cdot)$ denote the space of probability distributions over the argument. A finite horizon POMDP can be represented as tuple: $(\mathcal{S}, \mathcal{A}, T, R, \mathcal{Z}, O, b_0, H, \gamma)$, where $\mathcal{S}$ are the states, $\mathcal{A}$ are the actions, and $\mathcal{Z}$ are the observations. At planning iteration $t$, the agent selects an action $a \in \mathcal{A}$ and the transition function $T: \mathcal{S} \times \mathcal{A} \to \Pi(\mathcal{S})$ defines the probability of transitioning between states in the world, given the current state $s$ and control action $a$. After the state transition, the agent receives an observation according to the observation function $O: \mathcal{S} \times \mathcal{A} \to \Pi(\mathcal{Z})$, which defines the probability of receiving an observation, given the current state $s$ and previous control action $a$. The reward function $R: \mathcal{S} \times \mathcal{A} \to \mathbb{R}$ serves as a specification of the task. A POMDP is initialized with belief $b_0$ and plans over horizon $H$ with discount factor $\gamma$.  

Due to the stochastic, partially observable nature of current and future states, the realized reward in a POMDP is a random variable. Optimal planning is defined as finding a policy $\{\pi_t^*: \Pi(\mathcal{S}) \to \mathcal{A}\}_{t=0}^{H-1}$ that maximizes expected reward: $\mathbb{E} \Big[ \sum_{t=0}^{H-1} \gamma^t R\big(S_t, \pi_t(b_t)\big) \mid b_0 \Big]$, where $b_t$ is the updated belief at time $t$, conditioned on the history of actions and observations. The recursively defined horizon-$h$ optimal value function $V^*_h$ quantifies, for any belief $b$, the expected cumulative reward of following an optimal policy over the remaining planning iterations: $V_0^{*}(b) = \truemax_{a \in \mathcal{A}} \mathbb{E}_{s \sim b}[R(s, a)]$ and 
\begin{align}
     V_h^{*}(b) &=  \max_{a \in \mathcal{A}} \mathbb{E}_{s \sim b}[R(s, a)] + \gamma \int_{z \in \mathcal{Z}} P(z \mid b, a) V_{h-1}^*(b^{a,z}) \text{d}z \hspace{0.6cm} h \in [1, H-1],  \label{eq:value}
\end{align}
where $b^{a,z}$ is the updated belief after taking control action $a$ and receiving observation $z$, computed via Bayes rule using the transition $T$ and observation $O$ functions. The optimal policy at horizon $h$ is to act greedily according to a one-step look ahead of the horizon-$h$ value function. However, Eq.~\ref{eq:value} is intractable for large or continuous state, action, or observation spaces and thus the optimal policy must be approximated. Well-designed algorithmic and heuristic choices are necessary to develop efficient, practical, and robust planning algorithms for expeditionary sciences.

\section{Advances to Expeditionary Science}
\label{sec:fieldops}
\begin{figure}[h!]
    \centering
    \includegraphics[width=\columnwidth]{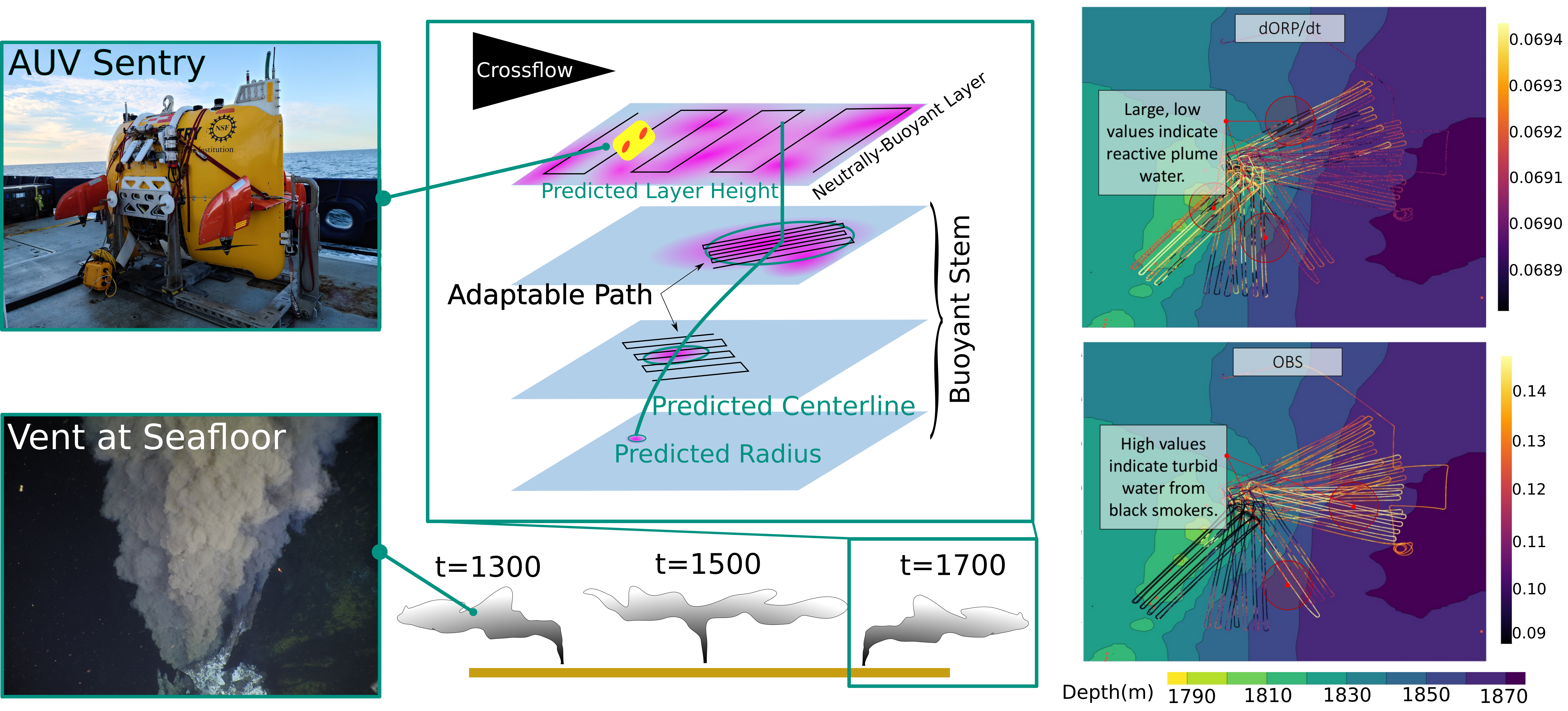}
    \caption{In November 2021, AUV Sentry was deployed in the Gulf of California to detect and map hydrothermal plume structures in the deep ocean. Plumes are composed of a buoyant stem and neutrally-buoyant layer. These plumes are dynamic, subject to tidal advection, diffusion, and turbulent mixing. We infer the spatiotemporal structure of plumes from sparse partial-observations and plan informative trajectories that respect operational constraints of AUV Sentry. Field data are shown on the right from two onboard science point-sensors. The dORP/dt sensor shows areas of reactive water, likely indicating buoyant stem detections. The OBS sensor shows areas of high turbidity, which can persist throughout a plume. Combining these sensor signals, we can infer the plume structure and dynamics.}
    \label{fig:diagram}
\end{figure}

As a case study we highlight a scientific mission to the Gulf of California in November 2021, in which a research vessel deployed an autonomous underwater vehicle (AUV) for deep sea discovery and mapping of hydrothermal vent plumes (see Fig.~\ref{fig:diagram}). Characterizing hydrothermalism has implications for global nutrient and energy budgets, but nearly two-thirds of all hypothesized hydrothermal vents are yet undiscovered \cite{beaulieu2015undiscovered}. AUVs with point sensors and the ability to execute trajectories under operational constraints can ``sniff'' for vents in the water column and use observations to estimate seafloor vent characteristics, but tidal advection, diffusion, and turbulent mixing all contribute to complicated spatiotemporal observations. Additionally, accurate vent inference requires the disambiguation of the spatially-small, chemically-rich buoyant stem from the spatially-vast neutrally-buoyant layer that compose plumes \cite{morton1956turbulent}. Here, we present some of our work tackling this expeditionary robotics problem, and highlight key challenges we found in this context.

\subsection{Spatiotemporal IPP for the Maximum Seek-and-Sample Problem}
We formulated the problem of finding the buoyant stem within the multi-modal non-buoyant plume layer using a robot as a POMDP and developed \PLUMES: \textbf{P}lume \textbf{L}ocalization under \textbf{U}ncertainty using \textbf{M}aximum-valu\textbf{E} information and \textbf{S}earch \cite{flaspohler2019information}, an autonomy stack for target-seeking in multi-modal environments. \PLUMES utilized a Gaussian Process (GP) belief to model a single time-varying chemical signal, a progressively-widened Monte Carlo tree search, and a novel task-driven information-theoretic reward function. In spatiotemporal systems, \PLUMES was able to find and track moving buoyant-stem targets and repeatedly explore an environment as uncertainty grew under stochastic plume dynamics \cite{preston2019adaptive}. \PLUMES addressed challenges in continuous search for planning and balancing exploration-exploitation behavior under sparse reward signals, and demonstrated how the sum of algorithmic components can lead to a potent system. However, \PLUMES also revealed the challenges of expeditionary science, particularly the difficulty of simultaneously training and utilizing a spatiotemporal belief model, and developing computationally tractable real-time nonmyopic planners. 

\subsection{Abstraction via Task-driven Macro-action Discovery}
One key barrier to tractable, real-time nonmyopic planning in \PLUMES was the need to search over long planning horizons, chaining together low-level action primitives to explore the environment. The development of abstract, high-level action representations would enable planners to search over shorter planning horizons, saving compute effort during real-time planning. Instead of using human-designed abstractions of the robot's continuous action space, our recent work learns a set of high-level actions that are useful for a given scientific task directly from a low-level specification of the POMDP model \cite{flaspohler2020belief}. The algorithm takes into account the robot's current uncertainty about the state of the world, the stochasticity of the underlying environmental dynamics, and the current task to develop a set of high-level actions that can be used online while providing formal performance guarantees about the resulting ``abstract'' policies. 

\subsection{Iterative Mission Structures and Physically-Informed Uncertainty Models}
\PLUMES used a general, data-driven model of the environment and assumed that the robot's action space consisted of flexible, low-level motion primitives. To improve the operational effectiveness of \PLUMES at sea, we proposed to: 1) make use of data-efficient probabilistic models that explicitly use physical equations that govern plume expression in the water column, and 2) develop planners that directly encode the physical and operational constraints of the AUV. We developed and tested this new algorithmic expeditionary system during field trials in November 2021. The system made use of a belief model called \PHUMES: \textbf{PH}ysically-informed \textbf{U}ncertainty \textbf{M}odel of \textbf{E}nvironment \textbf{S}patiotemporality, which sampled uncertain parameters of a simplified model of buoyancy-driven hydrothermalism (e.g., vent temperature, tidal crossflow) to generate multiple ``envelopes'' of possible plume-derived waters. These were aggregated into a probabilistic forecast over the spatiotemporal volume to be sampled by the AUV. A trajectory optimizer with knowledge of vehicle and ship operational constraints then scaled lawnmower primitives to maximize the number of ``in plume'' measurements that were to be collected at the continuum between buoyant stem and neutrally-buoyant layer. \PHUMES addressed open challenges in robotic decision-making related to injection of prior knowledge or scientific inductive bias for data efficient training in complex domains and scalable planning with expensive belief models.

\section{Core Challenges and Opportunities in Expeditionary Robotics}
\label{sec:challenges}
As identified in Sec.~\ref{sec:fieldops}, there are a number of open challenges in reinforcement learning and decision-making in the context of expeditionary robotics. Here we present a brief survey of these open challenges and opportunities informed by years of collaborating with scientists to develop expedition frameworks in diverse application areas.

\subsection{Belief Representations}


\textbf{Epistemic and aleatoric uncertainty:}
Reducing epistemic uncertainty of a spatiotemporal environment requires access to a model of the underlying dynamical system, or a data-driven technique that can uncover it. Extracting physically-meaningful quantities from observational data is typically performed post-expedition using computationally expensive numerical models ``tuned'' by observations. While this lends itself well to Bayesian inference formulations, it is intractable for practical decision-making. Data-driven techniques for model discovery \cite{raissi2019physics} may be arguably more tractable, but generally suffer small-data challenges. Developing models that overcome the challenges of efficiently characterizing spatiotemporal dynamics from streaming, sparse observations would generally improve expeditionary robotics. Additionally, there is a unique opportunity to enable computation of proxies for aleatoric uncertainty, which are well-described in spatiotemporal environments with measures of chaotic motion (e.g., Lyapunov exponents) inferred from data \cite{blanchard2019analytical}. The implication that aleatoric uncertainty can be estimated has yet to be utilized to, e.g., assess the attainable resolution of a model or set planning horizons.

\textbf{Scientific knowledge as inductive bias:}
The kernel of a GP, the loss function in a neural network, or the activation functions between layers in a deep network can all be viewed as forms of inductive bias in a learning problem. For data-driven discovery of spatiotemporal dynamics, improving sample efficiency by leveraging opportunities to inject scientific knowledge to alleviate the learning burden is an open problem. While canonical numerical models of spatiotemporal phenomena are too computationally expensive to directly incorporate into e.g., GP kernels, the physical principles that underlie these models can be more easily summarized. ``Physically-informed'' data-driven probabilistic representations have been demonstrated outside of expeditionary robotics \cite{raissi2019physics} and some work within IPP \cite{salam2019adaptive} shows rich opportunities for analyzing and extending these methods for larger environments and longer planning horizons.

\textbf{Low-dimensional state embeddings:}
Expressing a spatiotemporal environment completely would require an exceedingly large, high-dimensional representation. Model order reduction (MOR) techniques reduce the dimensionality of spatiotemporal systems to a set of weights and vectors that sufficiently describe patterns in the dynamics. Uncovering low-dimensional state embeddings from partially-observed expedition data is a general challenge; uncovering a \emph{useful} embedding for a specific decision-making problem is additionally challenging. Access to such an embedding would reduce the computational burden of representing belief in large environments for planning.


\subsection{Decision-Making}

\textbf{Rollout-based planning with expensive belief models:}
State-of-the-art planners for POMDP problems often make use of rollout-based planning in tree search frameworks; continuous search variables are handled using strategies such as progressive widening or scenario sampling \cite{sunberg2018online}. However, these planners require extensive online simulations for each rollout performed. Forward-simulating the dynamics and observational models for complex, spatiotemporal phenomena can be computationally intensive, which often limits the feasible look-ahead horizon in real-time operations on computationally-limited robotic platforms.  Planners that selectively or adaptively perform expensive rollouts, automatically adjust the planning horizon based on the dynamics of the environmental system, or make use of continuous, offline planners would enable improved decision-making for expeditionary science.


\textbf{Information rewards and task-driven exploration:} Due to partial observability and stochastic dynamics in spatiotemporal contexts, a decision-maker must operate with significant and often growing state uncertainty. However, not all state uncertainty impacts task performance and uniform information gathering strategies can be inefficient. Understanding the value of information for accomplishing a task is a known challenge for planning under uncertainty and this is particularly true for expeditionary robotics. Recent works that develop heuristic information rewards \cite{flaspohler2019information} or task-driven value of information metrics \cite{flaspohler2020belief} begin to build the tools necessary for expeditionary robotic planning.

\textbf{Robust planning under model mismatch and uncertainty:}
Scientific models, whether data-driven or based on physical principles, are always imperfect representations of a robot's environment. Model mismatch or uncertainty in key model parameters leads to discrepancies between the environmental predictions that a robot uses during planning and its real-time observations. Planning robot trajectories that entirely miss a phenomenon due to overconfidence in an incorrect model is detrimental to scientific objectives. Planners must develop policies or trajectories that are robust to model mismatch and uncertainty, or are guaranteed to perform as well as a simple, naive data collection strategy. 

\textbf{Interpretable and operational decision-making:}
Decision-making algorithms must interface with and are constrained by a variety of stakeholders, including scientists, robot operators, and engineers. For example, when deploying an AUV from an oceanographic research vessel, the decision-making algorithm must account for ship scheduling, timing delays, weather, and multi-vehicle operations. This requires developing flexible planners that can understand and account for these complex constraints. Additionally, stakeholders are often concerned with robot safety and data quality. Producing plans that are interpretable for scientific and operational stakeholders is key for building trust and confidence in scientific autonomy.

\section{Final Thoughts}
Expeditionary science and robotics motivates a set of interesting reinforcement learning and decision-making problems that are not well-addressed by current state-of-the-art methods. There is increasing need to enable better \emph{in situ} methodologies for monitoring and characterizing large spatiotemporal environments as accelerated climate change transforms delicate carbon budgets, ecosystem networks, and weather patterns. Robotic technologies can play a transformative role in understanding these evolving trends and assessing policy interventions. Novel developments towards the challenges listed here may have constructive and complementary effects in other application spaces, as many domains, such as material discovery and robotic manipulation, have overlapping challenges and opportunities. Our contributions, including \PLUMES \cite{flaspohler2019information,preston2019adaptive}, macro-action discovery \cite{flaspohler2020belief}, the \PHUMES model and trajectory optimizer for operational missions, and ongoing work in physically-informed deep kernel learning, represent algorithmic and systems contributions towards more effective autonomous systems in expeditionary science and begin to address some of the key challenges that we propose. 

\bibliographystyle{IEEEtran}
\bibliography{rldm}

\end{document}